\documentclass[10pt,twocolumn,letterpaper]{article}
\usepackage{pgfplots}
\usepackage{tikz}
\usepackage{iccv}
\usepackage{times}
\usepackage{epsfig}
\usepackage{graphicx}
\usepackage{amsmath}
\usepackage{amssymb}
\usepackage{booktabs}
\usepackage{verbatim}
\usepackage{color, colortbl}
\usepackage{pifont}
\usepackage{dsfont}
\usepackage{multirow}
\usepackage{pythonhighlight}
\usepackage[ruled,linesnumbered]{algorithm2e}
\usepackage[accsupp]{axessibility}

\usepackage[pagebackref=true,breaklinks=true,letterpaper=true,colorlinks,bookmarks=false]{hyperref}

\definecolor{beaublue}{rgb}{0.74, 0.83, 0.9}
\definecolor{palecornflowerblue}{rgb}{0.67, 0.8, 0.94}
\definecolor{lightblue}{rgb}{0.9, 0.9, 1.0}
\definecolor{lightgreen}{rgb}{0.92, 0.97, 0.87}
\definecolor{lightred}{rgb}{1.0, 0.85, 0.85}
\newcommand{\cmark}{\ding{51}}%
\newcommand{\xmark}{\ding{55}}%
\newcommand\minisection[1]{\vspace{1mm}\noindent \textbf{#1}}
\newcommand{\red}[1]{\textcolor{red}{#1}}

\definecolor{Gray}{gray}{0.9}
\newcommand{\model}{PivotNet\xspace}
\newcommand{\modelit}{\textit{PivotNet}\xspace}
\newcommand{\modelbf}{\textit{\textbf{PivotNet}}\xspace}
\newcommand{\bev}{\textit{BEV}\xspace}
\newcommand{\LD}{$lane$-$divider$\xspace}
\newcommand{\PC}{$ped$-$crossing$\xspace}
\newcommand{\RB}{$road$-$boundary$\xspace}
\newcommand{\hdmap}{\textit{HD map}\xspace}

\newcommand{\nuscenes}{\textit{nuScenes}\xspace}
\newcommand{\argoverse}{\textit{Argoverse 2}\xspace}

\usepackage[font=small,labelfont=small]{caption}
\usepackage[capitalize]{cleveref}
\usepackage{authblk}

\iccvfinalcopy 

\def\iccvPaperID{2536} 

\ificcvfinal\fi

\begin{document}

\title{\model: Vectorized Pivot Learning for End-to-end HD Map Construction}
\author{
	Wenjie Ding{\textsuperscript{\textasteriskcentered}} \quad Limeng Qiao\thanks{Equal Contribution} \quad Xi Qiu\thanks{Corresponding Author}  \quad Chi Zhang \\
	{MEGVII Technology} \\
	{\tt\small \{dingwenjie, qiaolimeng, qiuxi, zhangchi\}@megvii.com} 
	
}   

 


\maketitle
\ificcvfinal\thispagestyle{empty}\fi

\begin{abstract}
	Vectorized high-definition map online construction has garnered considerable attention in the field of autonomous driving research. Most existing approaches model changeable map elements using a fixed number of points, or predict local maps in a two-stage autoregressive manner, which may miss essential details and lead to error accumulation. {Towards precise map element learning, we propose a simple yet effective architecture named \modelbf, which adopts unified pivot-based map representations and is formulated as a direct set prediction paradigm.} Concretely, we first propose a novel {Point-to-Line Mask} module to encode both the subordinate and geometrical point-line priors in the network. Then, a well-designed Pivot Dynamic Matching module is proposed to model the topology in dynamic point sequences by introducing the concept of sequence matching. Furthermore, to supervise the position and topology of the vectorized point predictions, we propose a Dynamic Vectorized Sequence loss. Extensive experiments and ablations show that \model is remarkably superior to other SOTAs by 5.9 mAP at least. The code will be available soon.	
\end{abstract}

\section{Introduction}
High-definition map~(\hdmap) is one of the most critical components in many autonomous driving modules, including simulation, localization, and planning. Typical \hdmap construction relies on manual annotation on lidar point clouds, which is time-consuming and labor-intensive.~Recent works explore the map learning problems to reduce the labeling costs~\cite{maplearning1,li2022hdmapnet,maplearning2,vpn,cycletransform}. Given data from onboard sensors, map learning aims to construct local map within a predefined bird's-eye-view~(\bev) range.  

{Most existing works view the map construction as a semantic learning problem~\cite{maplearning1,maplearning2, vpn, cycletransform}. 
They represent a map within certain range as an evenly spaced field and predict the class label for each grid, generating a rasterized map. However, there are obvious limitations of rasterized representation in map learning. First, rasterized maps are composed of dense semantic pixels that contain redundant information, requiring large amounts of memory and transmission bandwidth, especially if the map extent is large. Second, the rasterized representation assumes the independence of map grids, which ignores the geometric relationship between and within map elements. Third, complex post-processing~\cite{li2022hdmapnet} is required to obtain vectorized maps for downstream tasks, which brings additional computation, time consumption, and accumulated errors. }

\begin{figure}[t]
	\centering
        \setlength{\belowcaptionskip}{-2pt}
	\includegraphics[width=8cm]{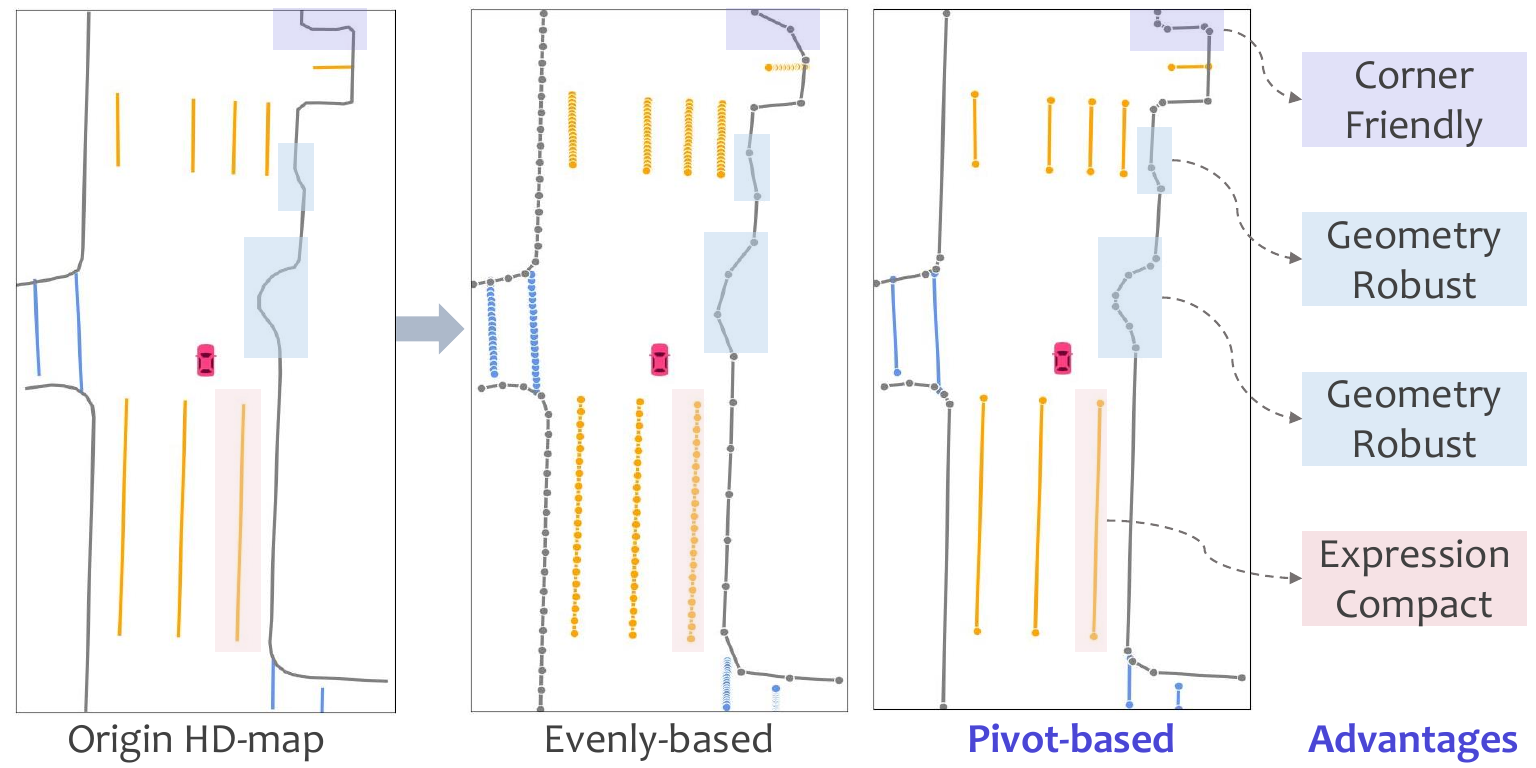} \\
	\caption{
		Illustration of our motivation for pivot-based vectorization.
		Given an original \hdmap example from \nuscenes, there are two different ways to construct a vectorized map, \ie \textit{evenly-based} and \textit{\textbf{pivot-based}}. 
		Comparison shows that pivot-based vectorization is more corner-friendly, geometry-robust, and expression-compact.
		Circles on the lines denote the real vectorization GT.
	}
\label{fig:diffmap}
\end{figure}
To address the limitations of current semantic learning methods, there have been proposals for generating vectorized representations in an end-to-end manner. One such method is MapTR~\cite{maptr}, which uses a fixed number of points to represent a map element, regardless of its shape complexity. However, this approach has two drawbacks. First, the evenly-based representation contains redundant points that have little effect on the geometry. Second, representing a dynamically shaped line with a fixed number of points may miss essential details in map elements, resulting in information loss, particularly for rounded corners and right angles~(Fig.\ref{fig:diffmap}). Therefore, to learn an accurate and compact representation, we model a map element as an ordered list of pivotal points, which is \textit{expression compact}, \textit{corner friendly}, and \textit{geometry robust}. However, the pivot-based representation brings new challenges due to the dynamic number of the pivot points within different map elements. {Previous work~\cite{vectormapnet} has utilized a coarse-to-fine framework and autoregressive decoding to address these challenges, but the autoregressive nature can lead to long inference time and accumulated errors.} Towards these issues, we propose \modelbf, which accurately models map elements through pivot-based representation in a set prediction framework. 

The framework of \model is depicted in Fig.~\ref{fig:overview}, which presents an architecture consisting of four primary modules: the camera feature extractor, \bev feature decoder, line-aware point decoder, and pivotal point predictor. The surrounding multi-view images from onboard cameras are fed into the camera feature extractor, which generates camera view features. Next, the image features are aggregated and transformed into a unified \bev feature through the \bev feature decoder. The lane-aware point decoder then extracts line-aware point features. Finally, the pivotal point predictor removes {collinear points} and predicts a flexible yet compact pivot-based representation.

To be concrete, we first propose a \textit{point-to-line mask module} for the line-aware point decoder, which encodes both the subordinate and geometric point-line relation through a line-aware mask. Secondly, we further design a \textit{pivot dynamic matching} module, which models the connection in pivotal point sequences by introducing the concept of sequence matching. A custom sequence matching algorithm is further devised to enhance the time efficiency. Lastly, we propose a novel \textit{dynamic vectorized sequence} loss to supervise the position and topology of the vectorized point predictions, through both pivot and collinear point supervision. By formulating the task as a sparse set prediction problem and leveraging an end-to-end sequence matching based bipartite matching loss, we present a method that generates precise yet compact vectorized representations without requiring any post-processing.

The contributions of the paper are threefold:
\begin{itemize}
	\item We present \modelbf, an end-to-end framework for precise yet compact \hdmap construction via pivot-based vectorization.
	\item We innovatively introduce \textit{point-to-line mask module}, \textit{pivot dynamic matching module}, and \textit{dynamic vectorized sequence} loss for accurate map element modeling.
	\item \model exhibits remarkable superiority over state of the arts (SOTAs) on existing benchmarks, indicating the effectiveness of our approach.
\end{itemize}

\label{sec:intro}

\section{Related Works}
\subsection{Semantic map learning} 
HD map encompasses intricate details that transcend the scope of standard maps, which amplifies the challenge of precise annotations. 
The conventional process of map construction hinges upon the utilization of LiDAR sensors. This intricate pipeline encompasses stages such as data collection, point cloud registration~\cite{lee2013robust,mendes2016icp,lego,liosam,lidarmap,semanticalign} and
manual annotation.
To curtail the labeling expenses and enhance overall efficiency, map learning techniques have been introduced. These methods aim to extract pertinent map elements from various on-board sensors, such as cameras and LiDAR sensors~\cite{maplearning1,maplearning2,li2022hdmapnet}. Most approaches generates semantic \bev map representations only. To transform the image features to the \bev space, VPN~\cite{vpn} utilize a multilayer perceptron to learn the mapping between camera views and the \bev. LSS~\cite{lss} and BEVDet~\cite{huang2021bevdet} bridge the view gap based on the depth distribution estimation. With the prevalent DETR~\cite{detr} paradigm, recent methods adopt \bev queries and encode the geometry prior in the attention mechanism in Transformer~\cite{persformer, bevformer,cycletransform,detr3d}. To obtain vectorized map for downstream tasks, HDMapNet~\cite{li2022hdmapnet} first produce semantic map and then groups pixel-wise segmentation results in the post-processing. Instead of adopting the semantic first and vectorization later pipeline, \model learns the vectorized map representation in an end-to-end manner. 

\subsection{Vectorized HD Map Construction} 
To avoid time-consuming post-processing, recent works explore vectorized map learning methods~\cite{maptr, vectormapnet} to obtain compact vectorized map in an end-to-end manner. It is challenging to model the topology between and within map elements due to the various geometric shape and complexity. MapTR~\cite{maptr} models a map element using a fixed number of points, which results in information loss, especially for the rounded corners and the right-angles. VectorMapNet~\cite{vectormapnet} utilizes a coarse-to-fine architecture with an autoregressive network, which leads to long inference time and possible accumulated errors. Different from the existing works~\cite{maptr, vectormapnet}, \model models a map element using a dynamic number of pivotal points and adopts a set prediction paradigm, which preserves map details while avoiding the drawbacks of \cite{vectormapnet}.  

\begin{figure*}[ht]
	\centering
	\includegraphics[width=17cm]{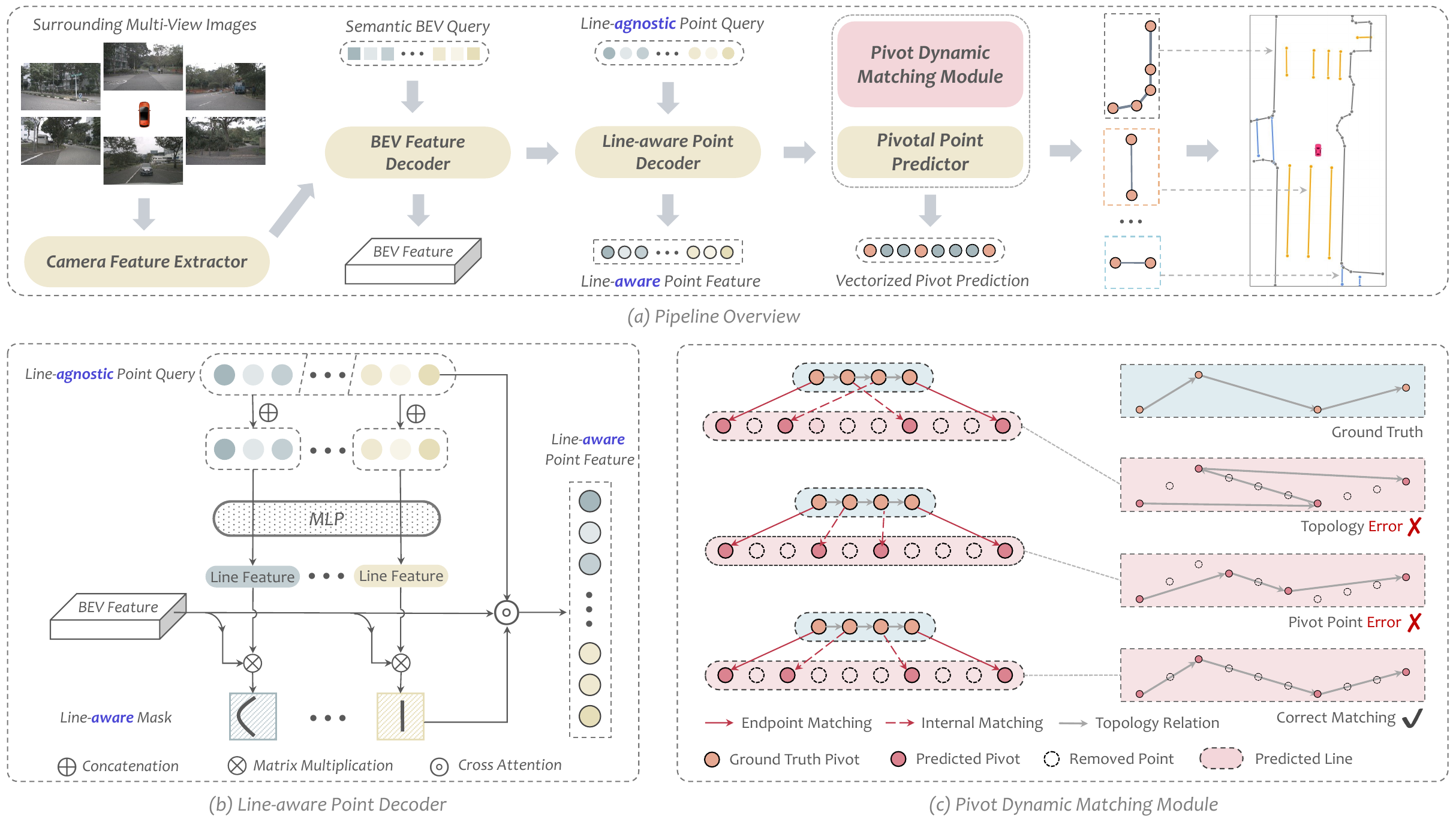} \\
	\caption{\textbf{An overview of the proposed \modelbf.} 
		The top row is the architecture pipeline of our model, 
		containing four primary components for extracting progressively richer information, 
		which takes the \textit{RGB} images as inputs and generates flexible and compact vectorized representation without any post-processing. The bottom row illustrates detailed structures of the \textit{line-aware point decoder} which decodes the map elements from the \bev feature, and the \textit{pivot dynamic matching module} which enables end-to-end sequence learning. 
	}
\label{fig:overview}
\end{figure*}

\subsection{Topology Modeling of Map Elements}

There are many works to model map elements in a sparse manner. Some methods generate map elements in a recurrent way. HDMapGen\cite{hdmapgen} generates synthetic \hdmap by using a hierarchical graph. Global graph with important points are first generated using a recurrent attention network and lane details are then generated using \textit{MLP}s. VectorMapNet~\cite{vectormapnet} adopts a coarse-to-fine scheme, which predicts coarse shape first and generates points on element in a recurrent way. The recurrent nature of these works makes them time-consuming and hard to train. Other methods formulate map element detection as a keypoint estimation and association problem~\cite{focusonlocal,wang2022keypoint}, which lack geometry relations modeling in point regression and need complex post-processing to group keypoints. Some anchor-based approaches~\cite{linecnn,laneatt} utilize the lane shape prior via special design on the anchor. There are also approaches that model the lanes as parameterized curves, such as polynomial curve~\cite{lstr, van2019end} and Bezier curve~\cite{feng2022rethinking,liu2020abcnet,Qiao_2023_CVPR}. Considering the changeful map elements, we choose the polyline representation. Rather than adopting the two-stage coarse-to-fine~\cite{vectormapnet,hdmapgen} or bottom-up~\cite{polyline_detect, focusonlocal,wang2022keypoint} design, we model both the point-level and line-level geometries in a uniform manner, and innovatively incorporate point-line prior through line-aware attention masks.

\section{Method}
We first present the formulation of vectorized \hdmap modeling with pivot-based representation in Sec.\ref{sec:formulation}. Then we elaborate on the design of \modelbf in Sec.\ref{sec:modeldesign}.
Lastly, we present the overall training loss in Sec.\ref{sec:end-to-end training}.

\subsection{Problem Formulation}
\label{sec:formulation}
\vspace{-0.2cm}

\begin{figure}[t]
	\centering
	\includegraphics[width=0.99\linewidth]{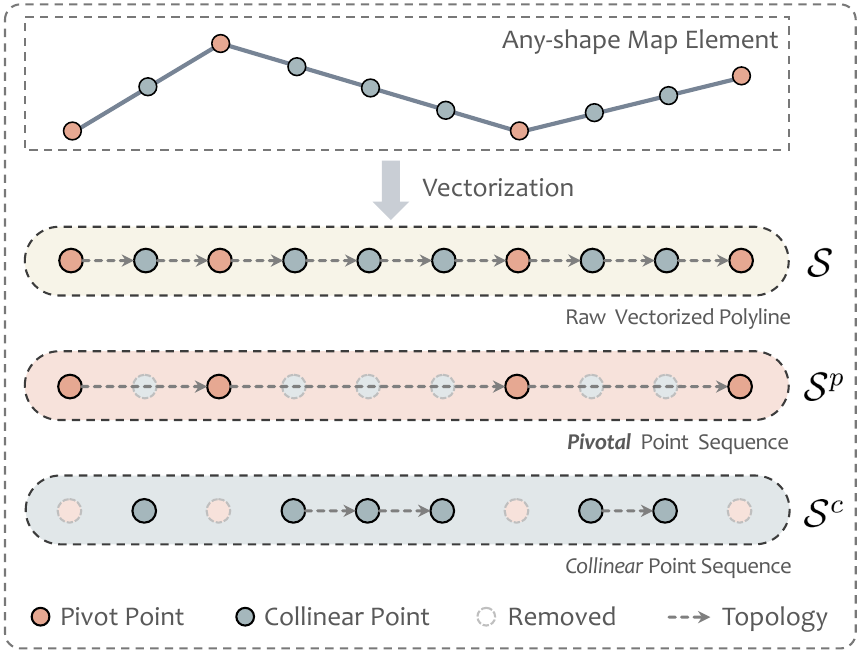} \\
	\caption{\textbf{Problem Formulation.} An any-shape map element is modeled as a vectorized sequence, which are split into pivotal sequence and collinear sequence. Pivot points are points that contribute to the overall shape and direction of the lines and are necessary to maintain its essential features. Collinear points refer to points that can be safely removed without affecting the line shape.} 
\label{fig:definition}
\end{figure}

Our objective is to generate vectorized representations for map elements in urban environments, utilizing data from onboard \textit{RGB} cameras~\cite{li2022hdmapnet, maptr, vectormapnet}. 
We illustrate the problem formulation in Fig.~\ref{fig:definition}. For each map element $\mathcal{L}$, we formulate it as a vectorized sequence $\mathcal{S}$ constructed by $N$ ordered points, where the connection of points is implicitly encoded in the index ordering. As $N$ tends to infinity, $\mathcal{S}$ tends to $\mathcal{L}$, which is formulated as $\displaystyle\lim_{N \to \infty}\mathcal{S}(N)=\mathcal{L}$. 
To form an efficient and compact representation, we further divide the vectorized sequence $\mathcal{S}$ into a pivotal point sequence $\mathcal{S}^p$ and a collinear point sequence $\mathcal{S}^c$ by the contribution to the map element shape. The pivotal sequence consists of a list of ordered pivot points that contribute to the overall shape and typically indicate a change in direction in a map element. Given a tolerable error $\epsilon$, $\mathcal{S}^p$ is supposed to be a subsequence of $\mathcal{S}$ that satisfies $d(\mathcal{S}^p, L) < \epsilon$ with the minimum sequence length, where $d(\cdot)$ denotes a distance metric. The collinear point sequence $\mathcal{S}^c$ is the complement sequence of $\mathcal{S}^p$ and $\mathcal{S}^c=\mathcal{S}-\mathcal{S}^p$. $\mathcal{S}^c$ consists of the points that are collinear to any two adjacent points in $\mathcal{S}^p$ and do not significantly contribute to the element shape. We denote the points in $\mathcal{S}^c$ as collinear points. 
Note map elements are often referred to as instances or lines in the next sections.

\minisection{Preliminary on vectorized \hdmap modeling.} 
We formulate the map construction task as a set prediction problem. We aim to learn a model that extracts compact information from the onboard cameras and predicts, for each map element, its corresponding pivotal point sequence, which uniquely determines the map element shape and position. Logically, there are three main challenges.

\noindent \textit{1)}~\textbf{\textit{Point-Line Relation Modeling.}} Based on the formulation, a map element is formed by a list of ordered points. For accurate modeling of map elements, it is crucial to encode the point-line relationship prior in the network. In {Sec.}\ref{sec:point-to-line-mask}, we propose the \textit{Point-to-Line Mask} module, which models both the subordinate and geometrical relation in a direct and interpretable manner.   

\noindent \textit{2)}~\textbf{\textit{Intra-instance Topology Modeling.}}
Vectorization necessitates precise topological relationships between points. For instance, {distinct topology with the same point set can represent entirely distinct line shapes} (refer to Fig.\ref{fig:overview}~(c)). In {Sec.\ref{sec:matching}}, we propose efficient \textit{Pivot Dynamic Matching} to model and constrain the topology within an instance.

\noindent \textit{3)}~\textbf{\textit{Intra-instance Dynamic Point Number Modeling.}} To achieve compactness in representation, our basic principle is to accurately represent map elements using the fewest possible pivotal points. Therefore, even for the same type of map element, such as a \RB, different instances require a dynamic number of points to achieve precise representation. {The proposed \textit{pivot dynamic matching} approach naturally solves the dynamic number problem through pivotal point classification.}

\subsection{\model}
\label{sec:modeldesign}
\subsubsection{Architecture Overview.}
\label{sec:overview}
The overall model architecture is presented in Fig.~\ref{fig:overview}, which consists of four primary components as follows:

\minisection{Camera Feature Extractor.}
Given multi-view images from the onboard cameras, a shared backbone network is adopted to extract image features. Then the multi-view image features are concatenated in order as outputs.  

\minisection{\bev Feature Decoder.} Following~\cite{persformer, bevformer}, we aggregate and transform the image features into a unified \bev representation via a Transformer-based module, while the deformable attention mechanism is adopted to reduce the computational memory. Specifically, a group of learnable parameters are predefined as \bev queries, where each query represents a grid cell in the \bev plane and only interacts with its regions of interest across camera views. To utilize the geometry prior between camera and \bev features, reference points of each \bev query are determined by the projection of \bev coordinates on the camera views.  

\minisection{Line-aware Point Decoder.}
We view the vectorized map construction as a set prediction task, and utilize a mask-attention based transformer to decode the lines from the \bev features. Specifically, this module takes the \bev features $F^b \in \mathbb{R}^{C\times H^b \times W^b}$ and a set of learnable point queries $\{Q_{m, n}\}_{m=1, n=1}^{M, N}$, where $Q_{m, n} \in \mathbb{R}^C$, and outputs a set of point descriptors $\{D_{m, n}\}_{m=1, n=1}^{M, N}$. Here $M$ is the max number of instances and $N$ is the max number of points in an instance. Each descriptor $D_{m, n} \in \mathbb{R}^C$ captures essential positional information and geometrical relationships within and between instances. Moreover, a descriptor $D_{m, n}$ corresponds to a point on the line, and an ordered list composed of descriptors $\{D_{m, n}\}_{n=1}^{N}$ represents a potential map element. To model the point-line relation, we propose a novel \textbf{\textit{Point-to-Line Mask}} module, which encodes both the subordinate and geometric relation in a straightforward manner with the added benefit of auxiliary supervision.

\minisection{Pivot Point Head.} This module consists of a \textit{pivotal point predictor} and a \textit{pivot dynamic matching module}. Given a set of point descriptors $\{D_{m, n}\}_{m=1,n=1}^{M,N}$, a set of point coordinates $\hat{V} = \{\hat{v}_{m,n}\}_{m=1,n=1}^{M,N}$ are first generated via point regression, where $\hat{v}_{m,n} \in \mathbb{R}^2$ denotes a point coordinate. {We define an ordered point sequence $\hat{\mathcal{S}}=\{\hat{v}_{m, n}\}_{n=1}^{N}$ to represent a map element with index $m$,} which contains both pivotal and collinear points. Therefore, to model the sequence order, we propose a novel \textit{\textbf{Pivot Dynamic Matching}} module for end-to-end sequence learning. Based on the matching results, $\hat{\mathcal{S}}$ is split into a pivotal sequence $\hat{\mathcal{S}}^p$ and a collinear sequence $\hat{\mathcal{S}}^c$. 
During inference, the \textit{pivotal point predictor} identifies the valid map elements and pivotal points, and outputs compact pivot-based representations.

\subsubsection{Point-Line Relation Modeling.}  
\label{sec:point-to-line-mask}
As illustrated in Sec.\ref{sec:overview}, we approach the vectorized map construction as a set prediction task. To accomplish this, we utilize a transformer network~\cite{mask2foremr} with each query representing a point. Yet, this approach poses an inherent challenge. Specifically, in the cross-attention module, there is no clear distinction between inter-instance and intra-instance points, which can result in mixed instances. Thus, it is crucial to encode the point-line relationship in the network to ensure accurate modeling of map elements. There exist both \textit{subordinate} and \textit{geometrical} priors between points and lines. The subordinate prior reflects the fact that some points belong to the same line, while others belong to different lines. The geometrical prior states that an ordered list of points contain the necessary information to form a line.

\minisection{Point-to-Line Mask Module}.
Previous research on map construction~\cite{maptr} has focused on encoding subordinate priors using hierarchical queries.
In this paper, we propose a novel module, called the \textit{\textbf{P}oint-to-\textbf{L}ine \textbf{M}ask} module~(PLM), that encodes both the subordinate and geometrical relations. The fundamental concept of this module is that a vectorized ordered point sequence is capable of constructing a map element. Based on this concept, we enforce the point queries of the same instance to learn a shared line-aware attention mask. The line mask is then incorporated in the cross-attention layer, implicitly encoding the subordinate relation through shared or different line-level masks. Additionally, the geometry relation is explicitly constrained through supervision on the line-aware mask.

As is shown in Fig.\ref{fig:overview}~(b), point queries $\{Q_{m, n}\}_{n=1}^{N}$ of the same instance with index $m$ are concatenated and fed into a multilayer perceptron, resulting in the instance-level line feature $I_m\in \mathbb{R}^C$. Then the line feature $I_m$ and the \bev feature map ${F}^b \in \mathbb{R}^{C\times H^b\times W^b}$ are multiplied to obtain the line-aware mask $\mathcal{M}_m \in \mathbb{R}^{H^b\times W^b}$. This mask is subsequently used in the cross-attention layer, along with the \bev feature and line-agnostic point queries, to produce line-aware point features. Notably, the line-aware mask effectively constrains the attention region of point queries to within the corresponding foreground map element region. Moreover, each unique attention mask is responsible for all the point queries of the same instance, which further enhances the subordinate prior encoding. 

\minisection{Line-aware Loss.} To ensure meaningful line features and constrain the geometrical relations, we introduce the line-aware loss~$\mathcal{L}_{LA}$, which is formulated as follows:
\begin{equation}
	\mathcal{L}_{LA} = \mathcal{L}_{bce}(\hat{M}_{line}, M_{line}) + \mathcal{L}_{dice}(\hat{M}_{line}, M_{line}).
\end{equation} 
Here $\hat{M}_{line}$ denotes the line-aware mask feature and $ M_{line}$ denotes the segmentation ground truth. $\mathcal{L}_{bce}$ and $\mathcal{L}_{dice}$ are the binary cross-entropy loss and dice loss~\cite{dice} respectively.

\subsubsection{Vectorized Pivot Learning.} 
\vspace{-0.2cm}
\label{sec:matching} 
 
\minisection{Pivot Dynamic Matching.} 
The {\textit{arbitrary shape}} and {\textit{dynamic point number}} of map elements bring challenges to topology modeling within instances. To address these issues, we propose \textit{{\textbf{P}ivot \textbf{D}ynamic \textbf{M}atching}}~(PDM), which models the connection in dynamic point sequence by introducing the concept of sequence matching. A custom matching algorithm is further proposed to enhance time efficiency.   

We consider the point matching problem between a predicted sequence $\hat{\mathcal{S}}=\{\hat{v}_{n}\}_{n=1}^{N}$ and a ground truth sequence $\mathcal{S}^p=\{{v}_n\}_{n=1}^{T}$. Here $N$ is the predefined max number of points in a line prediction and $T$ is the length of a ground truth sequence. $N$ is fixed while $T$ is dynamic depending on the map element shape. The instance index $m$ is omitted here for readability. $\hat{\mathcal{S}}$ contains both the pivot and collinear points, while $\mathcal{S}^p$ contains pivot points only. We denote a $T$-combination of the prediction $\hat{\mathcal{S}}$ sorted by point index as $\beta$. Apparently, if there are no constraints, the total number of unique $\beta$ is $C_N^{T}$. Examples of pivotal point matching are shown in Fig.\ref{fig:overview}~(c). Let's denote the $\beta(n)$-th point in the sequence prediction as $\hat{v}_{\beta(n)}$. Given $\hat{\mathcal{S}}$, $\mathcal{S}^p$ and a combination $\beta$, we define the sequence matching cost as,
\begin{equation}
\setlength{\abovedisplayskip}{1pt}
\setlength{\belowdisplayskip}{1pt}
	\mathcal{L}_{{match}}(\hat{\mathcal{S}}, \mathcal{S}^p, \beta) = \frac{1}{T}{\sum_{n=1}^{T} ||{v}_{n} - \hat{v}_{\beta^(n)}||_1},
\end{equation} 
where $||\cdot||_1$ denotes $L_1$ norm.
The proposed PDM searches for the optimal $\beta^{*}$ with the lowest sequence matching cost:  
\vspace{-0.15cm}
\begin{equation}
\setlength{\belowdisplayskip}{3pt}
    \beta^{*} = \mathop{\arg\min}\limits_{\beta} \mathcal{L}_{match}(\hat{\mathcal{S}}, \mathcal{S}^p, \beta),
\end{equation}
\vspace{-0.15cm}

Based on the matching result, $\hat{\mathcal{S}}$ is split into a pivot sequence $\hat{\mathcal{S}}^p$ and a collinear sequence $\hat{\mathcal{S}}^c$, where $\hat{\mathcal{S}}^p=\{\hat{v}_{\beta^{*}(n)}\}_{n=1}^T$ and $\hat{\mathcal{S}}^c=\hat{\mathcal{S}}-\hat{\mathcal{S}}^p$. For predicted lines with \textit{distinct} point distribution, the optimal $\beta^*$ is \textit{different}, resulting in distinct splits of pivot sequence $\hat{S}^p$ and collinear sequence $\hat{S}^c$. 
A brute force solution to find the optimal $\beta^*$ is to calculate the sequence matching cost for each $\beta$, leading to $O(C_N^{T})$ time complexity. To improve efficiency, we further devise a custom matching algorithm and reduce the time complexity to $O(NT)$.
As we treat the entire ordered sequence as a possible map element, a fixed correspondence of endpoints to those of the ground truth is enforced, \textit{i.e.}, $\beta(1) = 1$, $\beta(T) = N$. With such design, we are able to adopt the idea of dynamic programming. We use an array $dp$, where $dp[i][j]$ denotes the lowest sequence matching cost between the front-$i$ points in the target sequence and the front-$j$ points in the prediction sequences. Then, 
\begin{equation}
\setlength{\abovedisplayskip}{3pt}
\setlength{\belowdisplayskip}{3pt}
	dp[i][j] = \mathop{\min}\limits_{k \in [1, j-1] }dp[i-1][k]	+ ||v_i - \hat{v}_j||_1
\end{equation}
The base case is $dp[1][1] = ||v_1 - \hat{v}_1||_1$. The optimal $T$-combination $\beta^{*}$ is obtained during traversal in $dp$, which ends when $i=T,j=N$. We use another array to store the minimum cost during traversal to avoid unnecessary sorting, which is detailed in \textit{Supplementary Materials}.

\minisection{Dynamic Vectorized Sequence Loss.} 
To satisfy the problem formulation, we propose a novel \textit{\textbf{D}ynamic \textbf{V}ectorized \textbf{S}equence} loss~(DVS), which provides meaningful constraints for both the pivotal point sequence $\hat{\mathcal{S}}^p$ and the collinear sequence $\hat{\mathcal{S}}^c$, as well as the topology of the vectorized point predictions. DVS loss consists of three main parts, including {pivotal point supervision}, {collinear point supervision}, and {pivot classification loss}.

\noindent{1) \textit{Pivotal Point Supervision.}}
Based on the matching result, predicted pivot points in $\hat{S}^{p}$ is in one-to-one correspondence to the ground truth sequence $S^p$, and $|\hat{S}^{p}|=|S^p|=T$. Pivotal point loss constrains the $L_1$ distance between $\hat{\mathcal{S}}^p$ and the ground truth sequence $\mathcal{S}^p$, which is formulated as: 
\begin{equation}
	\mathcal{L}_{pp} = \frac{1}{T}{\sum_{n=1}^{T} ||\hat{\mathcal{S}}^p_{n} - \mathcal{S}^p_n||_1}.
\end{equation} 

\noindent{2) \textit{Collinear Point Supervision.}}
Based on the formulation, a collinear point in $\hat{\mathcal{S}}^c$ is supposed to be collinear to certain adjacent points in $\hat{\mathcal{S}}^p$ following the order in $\hat{\mathcal{S}}$. Assume there are $R_n$ collinear points between two adjacent pivot points $\hat{\mathcal{S}}^p_n$ and $\hat{\mathcal{S}}^p_{n+1}$, where $1\le n \le T-1$, then $|\hat{\mathcal{S}}^c| = \sum_{n=1}^{T-1}R_n=N-T$. We consider a collinear point $\hat{C}_{n,r}$ between $\hat{\mathcal{S}}^p_n$ and $\hat{\mathcal{S}}^p_{n+1}$ that ranks $r$ among $R_n$ collinear points. To ensure the \textit{linearity}, the target coordinate of $\hat{C}_{n,r}$ is supposed to be:
\begin{equation}
	C_{n,r} = (1-\theta_{n,r})S^p_n + \theta_{n,r}S^p_{n+1},
\end{equation} 
where $\theta_{n,r}$ is a coefficient that controls the relative position and $0<\theta_{n,r}<1$. Larger $\theta_{n,r}$ represents that $C_{n,r}$ is nearer to $S^p_{n+1}$ while farther from $S^p_n$. Therefore, to ensure the \textit{ordering} of the sequence prediction, $\theta_{n,r}$ is supposed to increase monotonically with $r$. For implementation convenience, we define $\theta_{n,r}=r/(R_n+1)$. Then the collinear point loss is formulated as:
\begin{equation}
	\mathcal{L}_{cp} = \frac{1}{N-T}\sum_{n=1}^{T-1}\sum_{r=1}^{R_n}||\hat{C}_{n,r} - C_{n,r}||_1.
\end{equation} 

\noindent{3) \textit{Pivot Classification Loss.}}
To model the dynamic pivotal point number, a binary cross-entropy loss is adopted to supervise the probability of a predicted point being a pivotal point. Given the probability of each point $\{p_n\}_{n=1}^{N}$ in an element prediction, classification loss is formulated as:
\begin{equation}
	\mathcal{L}_{cls} = \frac{1}{N}\sum_{n=1}^{N}\mathcal{L}_{bce}(p_n, \mathds{1}_{\hat{\mathcal{S}}_n \in \hat{\mathcal{S}}^p}),
\end{equation} 
where $N$ is the predefined maximum number of points in a map element. $\mathds{1}_A$ is an indicator function which returns $1$ if $A$ is true, and returns $0$ otherwise. 

Then the \textbf{\textit{DVS}} loss is formulated as follows:
\begin{equation}
	\mathcal{L}_{DVS} = \alpha_1\mathcal{L}_{pp} + \alpha_2\mathcal{L}_{cp}  + \alpha_3\mathcal{L}_{cls},
\end{equation} 
where $\alpha_1$, $\alpha_2$ and $\alpha_3$ denote the weighted factors.

\subsection{Training Loss}
\label{sec:end-to-end training}

\minisection{Auxiliary \textit{BEV} Supervision.} 
To ensure that \bev features contain necessary map information, we introduce an auxiliary segmentation-based loss for \bev supervision: 
\begin{equation}
	\mathcal{L}_{BEV} = \mathcal{L}_{bce}(\hat{M}_{bev}, M_{bev}) + \mathcal{L}_{dice}(\hat{M}_{bev}, M_{bev}).
\end{equation}  
$\hat{M}_{bev}$ is the predicted \bev mask and $M_{bev}$ is the ground truth. $\mathcal{L}_{bce}$ represents the binary cross-entropy loss and $\mathcal{L}_{dice}$ denotes dice loss~\cite{dice}.

\minisection{Overall Loss.} The overall loss is formulated as follows:
\begin{equation}
	\mathcal{L}_{TOTAL} = \mathcal{L}_{DVS} + \lambda_{1}\mathcal{L}_{LA} + \lambda_{2}\mathcal{L}_{BEV},
\end{equation}  
where $\lambda_1$ and $\lambda_2$ are weighted factors.

\begin{table*}[ht]
	\begin{center}
		\resizebox{1.0\textwidth}{!}{
			\begin{tabular}{ccc|cccc|cccc|cc}
				\toprule
				\multirow{2}{*}{Method} & \multirow{2}{*}{BKB} & \multirow{2}{*}{Epoch} & AP$_{\textit{divider}}$      & AP$_{\textit{ped}}$      & AP$_{\textit{boundary}}$      & mAP     & AP$_{\textit{divider}}$      & AP$_{\textit{ped}}$      & AP$_{\textit{boundary}}$      & mAP  & \multirow{2}{*}{FPS} & \multirow{2}{*}{Params.} \\ 
				\cline{4-11}
				&                      &                        & \multicolumn{4}{c|}{$\{0.2, 0.5, 1.0\}m$} & \multicolumn{4}{c|}{$\{0.5, 1.0, 1.5\}m$}    &                      &                       \\ 
				\midrule
				LSS~\cite{lss} & EB0 & 30 & 22.9 & 5.1 & 24.2 & 17.4 & - & - & - & - & - & - \\
				VPN~\cite{vpn} & EB0 & 30 & 22.1 & 5.2 & 25.3 & 17.5 & - & - & -  & - & - & - \\
				HDMapNet~\cite{li2022hdmapnet} & EB0 & 30 & 28.3 & 7.1 & 32.6 & 22.7 & - & - & - & - & - & -\\
				$^\ddagger$HDMapNet~\cite{li2022hdmapnet} & EB0 & 30 & 17.7$^\ddagger$ & 13.6$^\ddagger$ & 32.7$^\ddagger$ & 21.3$^\ddagger$ & 23.6$^\ddagger$ & 24.1$^\ddagger$& 43.5$^\ddagger$ & 31.4$^\ddagger$ & 0.7$^\ddagger$ & 69.8M \\
				VectorMapNet~\cite{vectormapnet} & R50 & 110 &27.2$^\dagger$ &  18.2$^\dagger$ & 18.4$^\dagger$ & 21.3$^\dagger$ &47.3 & 36.1 & 39.3 & 40.9 & 1.2$^\dagger$  & 19.4M \\
				MapTR~\cite{maptr} & R50 & 24 & 30.7$^\dagger$ & 23.2$^\dagger$ & 28.2$^\dagger$ & 27.3$^\dagger$ & 51.5 & 46.3 & 53.1 & 50.3 & \textbf{10.1}$^\dagger$  & 35.9M \\
				\midrule
				\rowcolor{lightgreen}
				\model (Ours) & EB0 & {24} &  39.4 &  32.9 &  37.1 & 36.5 & {55.7} & {55.1} & {58.5} & {56.4}  & 7.7 & 17.1M \\ 
				\rowcolor{lightgreen}
				\model (Ours) & R50 & {24} &  41.4 &  34.3 &  39.8 & 38.5 & {56.5} & {56.2} & {60.1} & {57.6}  & 6.7 & 41.2M \\ 
				\rowcolor{lightgreen}
				\model (Ours) & SwinT & {24} &  45.0 &  36.2 &  41.2 & 40.8 & {60.6} & {59.2} & {62.2} & {60.6}  & 5.1 & 44.8M \\ 
				\midrule
				\rowcolor{lightblue}
				\model (Ours) & EB0 & 30 & 43.7 & 34.7 & 40.2 & 39.6  & 59.7 & 53.9 & 61.0 & 58.2 & 7.7 & \textbf{17.1M} \\
				\rowcolor{lightblue}
				\model (Ours) & R50 & 30 & 42.9 & 34.8 & 39.3 & 39.0 & 58.8 & 53.8 & 59.6 & 57.4 &6.7 & 41.2M \\
				\rowcolor{lightblue}
				\model (Ours) & SwinT & 30 & 47.6 & 38.3 & 43.8 & 43.3 & 63.8 & 58.7 & 64.9 & 62.5 & 5.1 & 44.8M \\
				\rowcolor{lightblue}

				\model (Ours) & SwinT & 110 & \textbf{53.6} & \textbf{43.4} & \textbf{50.5} & \textbf{49.2} & \textbf{68.0} & \textbf{62.6}  & \textbf{69.7} & \textbf{66.8} & 5.1 & 44.8M \\

				\bottomrule
			\end{tabular}
			}
	\end{center}
\vspace*{-0.45cm}
\caption{
	Comparison with SOTAs on \nuscenes.  The $^\dagger$ indicates that results are re-evaluated on the tighter threshold setup with their released model checkpoint. And $^\ddagger$ represents that the performances are reproduced with their public codes. Results in \colorbox{lightgreen}{green} and  \colorbox{lightblue}{blue} shades mean that the pedestrian crossing are modeled with polygon as ~\cite{maptr,vectormapnet} and {straight lines} as ~\cite{li2022hdmapnet}. All mAP are obtained by averaging across three map elements. All numbers in the column of FPS  are re-test in the same $2080$Ti GPU device for fair comparison. 
}
\label{tab:main-result-nuscenes}
\end{table*}

\section{Experiments}
\vspace{-0.15cm}
\subsection{Experimental Settings}
\vspace{-0.25cm}
\minisection{Benchmarks.}
We evaluate \model on two popular and large-scale datasets, \ie \nuscenes~\cite{caesar2020nuscenes} and \argoverse~\cite{argoverse2}.
The \nuscenes dataset is annotated with $2$Hz. Each frame contains \textit{RGB} images from surrounding $6$ cameras, which cover full $360$ degree field of view. There are $1000$ scenes in the dataset, where each scene contains around $40$ frames. The dataset is split to $28K$ frames for training and $6K$ frames for validation.
The \argoverse is annotated with $10$Hz. Each frame contains $7$ ring cameras and $2$ stereo cameras. We use images from the ring cameras only. There are $110K$ frames for train and $24K$ frames for validation. 

\minisection{Evaluation Protocol.}
Following previous methods~\cite{li2022hdmapnet,maptr,vectormapnet}, we consider three categories of map element, namely \LD, \PC, and \RB for evaluating \hdmap construction.
Note the prediction map range is defined as $30m$ front and rear and $15m$ left and right of the vehicle.
The common average precision (AP) based on Chamfer Distance is adopted as the evaluation metric. We consider AP under three thresholds of $\{0.2, 0.5, 1.0\}m$ in default, where a prediction is treated as \textit{true positive (TP)} only if the distance between prediction and ground-truth is less than the specified threshold.
Furthermore, evaluation with an easier threshold setting of $\{0.5, 1.0, 1.5\}m$ is also taken into account in Table~\ref{tab:main-result-nuscenes} for fair comparison. 

\minisection{Implementation Details.}
We adopt EfficientNet-B0~\cite{tan2019efficientnet}, ResNet50~\cite{resnet}, and SwinTiny~\cite{liu2021swin} as backbones and employ transformer-based architecture as \bev feature extractor and line-aware point decoder, whose number of encoder/decoder layers are set to $4$/$4$ and $0$/$6$ respectively.
Moreover, we set the \bev feature size to $64 \times 32$ and the max number of point queries to $\{200, 50, 450\}$ for \LD, \PC, and \RB, where the max number of instances $M=\{20, 25, 15\}$ and the max number of points $N=\{10, 2, 30\}$.
Our model is trained on $8$ NVIDIA $2080$Ti GPUs with batch size $1$ per GPU. 
We use the AdamW~\cite{adamw} optimizer with learning rate $2e^{-4}$ and weight decay $1e^{-4}$.
Following the multistep scheduler, the learning rate is decayed by a factor of $0.2$ when training to the schedule of $0.7$ and $0.9$ of total epochs.
The weighted factors $\{\alpha_1,\alpha_2,\alpha_3,\lambda_1,\lambda_2\}$ are set to $\{5,2,2,5,3\}$ respectively without fine tune.

\begin{table}[t]
	\resizebox{0.48\textwidth}{!}{
		\begin{tabular}{ccc|cccc}
			\toprule
			Method & BKB & Epoch & AP$_{\textit{divider}}$ & AP$_{\textit{ped}}$ & AP$_{\textit{boundary}}$ &  mAP   \\
			\midrule
			$^\ddagger$HDMapNet~\cite{li2022hdmapnet} & EB0 & 6 & 19.5 & 9.8 & 35.9 &  21.8 \\
			$^\ddagger$VectorMapNet~\cite{vectormapnet} & R50 & 24 & 33.3 & 18.3 & 20.4 & 24.0 \\
			$^\ddagger$MapTR~\cite{maptr} & R50 & 6 & 42.2 & 28.3 & 33.7 & 34.8 \\
			\midrule
                \rowcolor{lightgreen}
                \model (Ours) & EB0 & 6 &  46.4 & {29.8} & 42.4 & 39.5  \\
                \rowcolor{lightgreen}
			\model (Ours) & R50 & 6 &  47.5 & {31.3} & 43.4 & 40.7  \\
                \rowcolor{lightgreen}
			\model (Ours)  & SwinT & 6 & {48.0} & {30.6} & {44.5} & {41.0} \\
                \rowcolor{lightgreen}
			\model (Ours)  & SwinT & 10 & \textbf{51.1} & \textbf{36.1} & \textbf{47.8} & \textbf{45.0} \\
			\bottomrule
		\end{tabular}
	}
\vspace{-0.25cm}
\caption{
	Comparison with state-of-the-art method on \argoverse.  The meanings of symbol $^\ddagger$ is the same as in Table \ref{tab:main-result-nuscenes}. 
}
\label{tab:main-result-argoverse}
\end{table}
\vspace{-0.15cm}
\subsection{Comparisons with State-of-the-art Methods}
\vspace{-0.2cm}
\minisection{Results on \nuscenes.}
In Table~\ref{tab:main-result-nuscenes}, we compare the overall evaluation performance of \modelit with existing \textit{SOTAs} on \nuscenes~\cite{caesar2020nuscenes} under different settings.
Existing methods use different AP thresholds in evaluation. Note \cite{li2022hdmapnet} employs the threshold of $\{0.2, 0.5, 1.0\}m$ while \cite{vectormapnet} and \cite{maptr} adopt an easier setting of $\{0.5, 1.0, 1.5\}m$. Therefore, we evaluate \modelit with both settings in Table~\ref{tab:main-result-nuscenes} for a fair comparison. Compared to the existing state-of-the-art ~\cite{maptr}, we achieve $11.2$ higher mAP with the $\{0.2, 0.5, 1.0\}m$ setting and ${7.3}$ higher mAP with the $\{0.5, 1.0, 1.5\}m$ setting. The results of \modelit with various backbones and training epochs are also provided. The reproduced performances of existing methods~\cite{li2022hdmapnet,maptr, vpn, lss} are obtained based on the public source code ($^{\ddagger}$) and released model checkpoint ($^{\dagger}$).

\minisection{Results on \argoverse.}
Table~\ref{tab:main-result-argoverse} benchmarks \modelit on a newly large-scale dataset \argoverse~\cite{argoverse2} and compares the performance with SOTAs.
The methods~\cite{li2022hdmapnet,maptr,vectormapnet} are reproduced with the public source code and then adapted to the \argoverse benchmark. 
Note all results in Table~\ref{tab:main-result-argoverse} are evaluated with the $\{0.2, 0.5, 1.0\}m$ threshold, and with pedestrian crossings modeled by polygons.
As can be seen, \modelit is superior to the existing SOTA approaches~\cite{li2022hdmapnet} by a considerable margin on \argoverse. 

\begin{table}[t]
	\resizebox{0.47\textwidth}{!}{
		\begin{tabular}{c|cc|cccc}
			\hline
			\# Row & PLM & PDM & AP$_{\textit{divider}}$ & AP$_{\textit{ped}}$ & AP$_{\textit{boundary}}$ &  mAP    \\
			\toprule
			1  & \xmark & \xmark &  33.8 & 36.8 & 32.4 & 34.4 \\
			2 & \cmark & \xmark &  42.3 & 35.5 & 34.9 & 37.6  \\
			3 & \xmark & \cmark & 40.6 & 37.4 & 39.6 & 39.2 \\
			4 & \cmark & \cmark & \textbf{47.6} & \textbf{38.3} & \textbf{43.8} & \textbf{43.3}\\
			\bottomrule
		\end{tabular}
	}
	\vspace{-0.15cm}
	\caption{Effectiveness of different modules in \modelbf. All results are conducted on \nuscenes with thresholds $\{0.2, 0.5, 1.0\}m$. PLM denotes the proposed \textit{point-to-line mask module} and PDM denotes the \textit{pivotal dynamic matching}.}
	\label{tab:Ablation-main-modules}
\end{table}

\subsection{Ablation Study}

\minisection{Effectiveness of different modules.}
We conduct ablations on \nuscenes to carefully analyze how much each module contributes to the final performance of \modelbf.
Table~\ref{tab:Ablation-main-modules} summarizes all results in great details. 
Specifically, the first row represents the baseline method, which adopts a vanilla transformer-based point decoder and a classification-based length predictor (as point sequence matcher).
As a simple alternative of sequence matching for arbitrary length sequence modeling, the later predicts the number of pivotal points as $k$ and outputs front-$k$ points as the final line.
{Comparison between row~2 and row~4 shows the effectiveness of dynamic sequence matching, especially on the complex-shaped \RB with $8.9\%$  higher AP.
Comparison between row~3 and row~4 validates the effectiveness of the proposed \textit{point-to-line mask module} with $4.9\%$ higher mAP.
Finally, we integrate the above two modules into baseline and show the final improvements in Row~4.}

\begin{table}[t]\footnotesize
	\resizebox{0.48\textwidth}{!}{
		\begin{tabular}{c|cccc}
			\hline
			$\mathcal{L}_{BEV}$ & AP$_{\textit{divider}}$ & AP$_{\textit{ped}}$ & AP$_{\textit{boundary}}$ &  mAP    \\
			\toprule
			\xmark &  44.3 & 37.4 & 42.4 & 41.4 \\
			\cmark & \textbf{47.6}\textbf{\scriptsize{\color{blue}(+3.3)}} & \textbf{38.3}\textbf{\scriptsize{\color{blue}(+0.9)}} & \textbf{43.8}\textbf{\scriptsize{\color{blue}(+1.4)}} & \textbf{43.3}\textbf{\scriptsize{\color{blue}(+1.9)}}  \\
			\bottomrule
		\end{tabular}
	}
	\vspace{-0.15cm}
	\caption{Effectiveness of \bev supervision. All results are conducted on \nuscenes with thresholds $\{0.2, 0.5, 1.0\}m$.}
	\label{tab:bev}
\end{table}

\begin{table}[t]
	\resizebox{0.48\textwidth}{!}{
		\begin{tabular}{c|cccc}
			\hline
			Method & AP$_{\textit{divider}}$ & AP$_{\textit{ped}}$ & AP$_{\textit{boundary}}$ &  mAP    \\
			\toprule
			Point Query &  40.6 & 37.4 & 39.6 & 39.2 \\
			Hierarchical Query & 37.4\textbf{\scriptsize{\color{red}(-3.2)}} & 33.1\textbf{\scriptsize{\color{red}(-4.3)}} & 37.3\textbf{\scriptsize{\color{red}(-2.3)}} & 35.9\textbf{\scriptsize{\color{red}(-3.3)}}  \\
			\midrule
			PLM~(Ours) & \textbf{47.6}\textbf{\scriptsize{\color{blue}(+7.0)}} & \textbf{38.3}\textbf{\scriptsize{\color{blue}(+0.9)}} & \textbf{43.8}\textbf{\scriptsize{\color{blue}(+4.2)}} & \textbf{43.3}\textbf{\scriptsize{\color{blue}(+4.1)}} \\
			\bottomrule
		\end{tabular}
	}
	\vspace{-0.2cm}
	\caption{
		{Comparison of \textit{point-line relation} modeling methods}.
			}
	\label{tab:Ablation-map-decoder-dwj}
\end{table}

\begin{figure}[t]
	\centering
	\includegraphics[width=6cm]{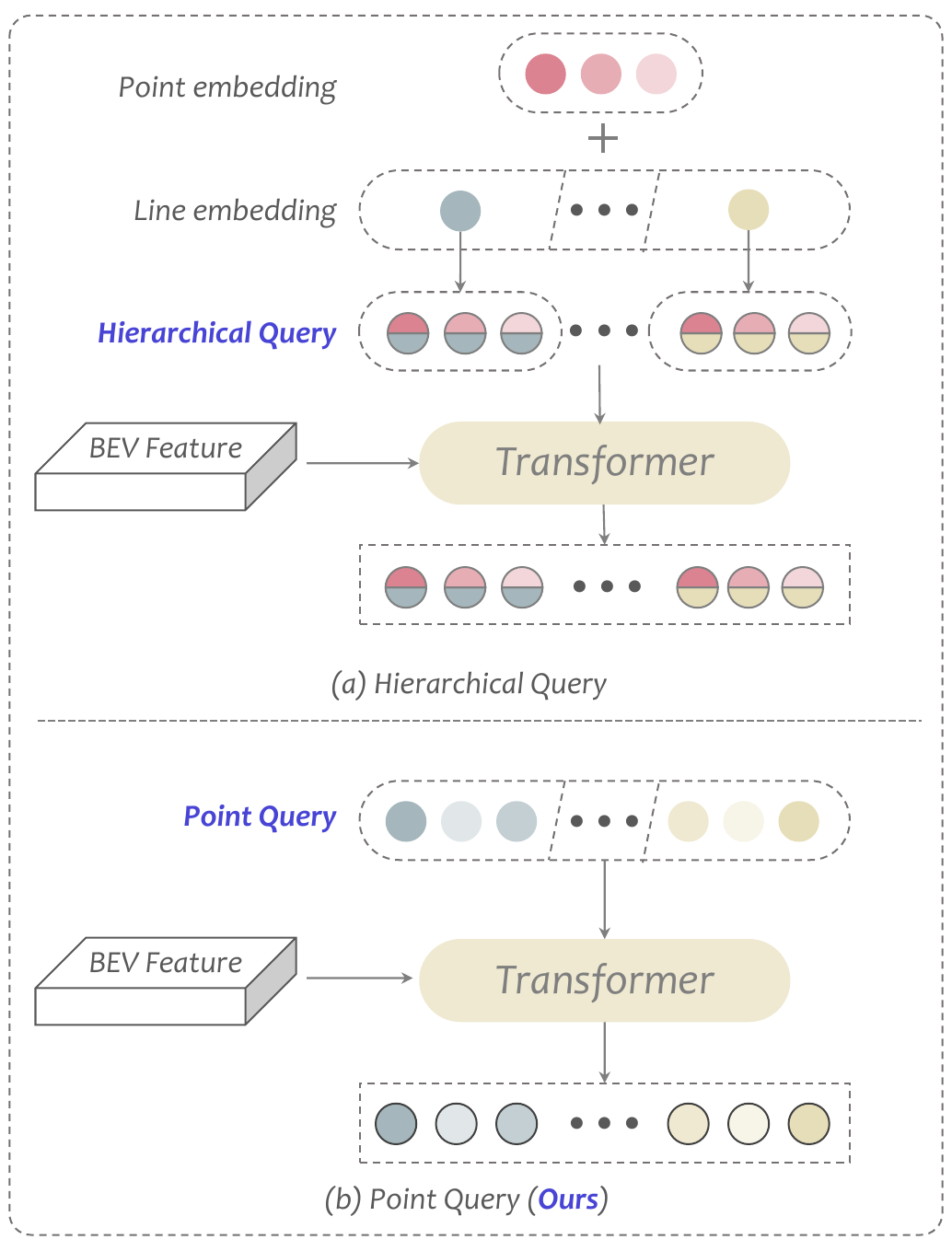} \\
	\caption{
		Comparison between hierarchical query and point query design. Hierarchical queries of the same line share the same line embedding and those with the same point index share the same point embedding. Point queries are both line-independent and index-independent. 
	}
\label{fig:hierarchical}
\end{figure}

\minisection{Effectiveness of \bev loss.} Table.~\ref{tab:bev} shows the effectiveness of \bev supervision. $\mathcal{L}_{bev}$ improves the performance by $1.9\%$ mAP by constraining the \bev features to capture meaningful information within the \bev range.

\minisection{Discussion of the point-line relation modeling.} In Table~\ref{tab:Ablation-map-decoder-dwj}, we compare different methods to model the point-line relation, including the \textit{point-to-line mask module}~(PLM) and the existing methods~\cite{maptr} with hierarchical queries. Row~1 represents the baseline of the line-aware point decoder with no point-line prior, utilizing a vallina mask-attention based transformer~\cite{mask2foremr}. Comparison between row 1 and 2 suggests that the hierarchical query design like \cite{maptr} is unhelpful to the dynamic sequence modeling. The reason is below. 
We present the hierarchical queries and the point queries in Fig.\ref{fig:hierarchical}. In the hierarchical query design~\cite{maptr}, point queries with the same index share the same point embedding. However, as we model a map element using a dynamic number of points, there is little relation between points with the same index in different line prediction. As illustrated in Sec.\ref{sec:matching}, the indexes of points selected to form the pivot sequences are various, depending on the point distribution. In short, the hierarchical embedding is \textit{{index-dependent}}, which is in conflict with our \textit{{index-independent}} problem formulation and may introduce noises about the index information. Therefore, to avoid the index-dependent embedding, we adopt point queries and embed the subordinate and geometrical point-line relation in the attention masks.  

\begin{table}[ht]
	\resizebox{0.48\textwidth}{!}{
		\begin{tabular}{c|cccc}
			\hline
			Method & mAP$_{0.2m}$ & mAP$_{0.5m}$ & mAP$_{1.0m}$ & mAP$_{1.5m}$   \\
			\toprule
			HDMapNet~\cite{li2022hdmapnet}  & 11.5 &  20.8 & 31.9 & 38.6 \\
			VectorMapNet~\cite{vectormapnet}  & 1.1\textbf{\scriptsize{\color{red}(-10.4)}} & 16.6\textbf{\scriptsize{\color{red}(-4.2)}} & 46.2\textbf{\scriptsize{\color{blue}(+14.3)}} & 64.0\textbf{\scriptsize{\color{blue}(+25.4)}} \\
			MapTR~\cite{maptr}  & 2.2\textbf{\scriptsize{\color{red}(-9.3)}} & 24.7\textbf{\scriptsize{\color{blue}(+3.9)}} & 55.1\textbf{\scriptsize{\color{blue}(+23.2)}} & 70.1\textbf{\scriptsize{\color{blue}(+31.5)}}  \\
			\midrule
			\model (Ours) & \textbf{13.3\textbf{\scriptsize{\color{blue}(+1.8)}}} & \textbf{40.8\textbf{\scriptsize{\color{blue}(+20.0)}} }& \textbf{61.4\textbf{\scriptsize{\color{blue}(+29.5)}}} & \textbf{70.6\textbf{\scriptsize{\color{blue}(+32.0)}}} \\
			\bottomrule
		\end{tabular}
	}
	\vspace{-0.25cm}
	\caption{Comparison of improvement under different thresholds. Note the mAP is obtained by averaging across three map elements.}
	\label{tab:vis-thre}
\end{table}

\minisection{Discussion of different AP thresholds.} 
Different threshold setups represent different tolerance degree of the evaluation protocol to model performance. Considering the application of auto-driving, \hdmap usually requires centimeter-level information, so we argue that the improvement under stricter threshold is more practical.  Compared with HDMapNet~\cite{li2022hdmapnet}, VectorMapNet~\cite{vectormapnet} and MapTR~\cite{maptr} significantly improves under simple thresholds, \eg $1.0m$, $1.5m$, but the improvement margin drop rapidly in stricter scenarios, \eg  $0.2m$, $0.5m$. Table \ref{tab:vis-thre} shows that no matter which threshold is token, our model always achieves better results, which shows the robustness of our framework.

\begin{table}[ht]\scriptsize
	\resizebox{0.48\textwidth}{!}{
		\begin{tabular}{c|cccc}
			\hline
			Layer Num. & AP$_{\textit{divider}}$ & AP$_{\textit{ped}}$ & AP$_{\textit{boundary}}$ &  mAP    \\
			\toprule
			1 &  37.8 & 34.4 & 36.1 & 36.1 \\
			3 &  46.7 & 36.8 & 41.9 & 41.8 \\
			\rowcolor{Gray}
			6 &  \textbf{47.6} & \textbf{38.3} & \textbf{43.8} & \textbf{43.3} \\
			7 &  47.1 & 37.7 & 42.8 & 42.6 \\
			\bottomrule
		\end{tabular}
	}
\vspace{-0.15cm}
\caption{Impact of line-aware point decoder layer number. The gray row represents the setting we use in default.}
\label{tab:Ablation-layer}
\end{table}
\vspace{-0.2cm}
\minisection{Impact of line-aware point decoder layer number.} The effect of the layer number in \textit{line-aware point decoder} is evaluated in Table~\ref{tab:Ablation-layer}. The performance of \model improves with more layer and saturates at layer number of $6$. Therefore, we stack $6$ layers for the decoder.

\begin{figure*}[htb]
	\centering
        \setlength{\abovedisplayskip}{0.cm}
	\includegraphics[width=0.98\linewidth]{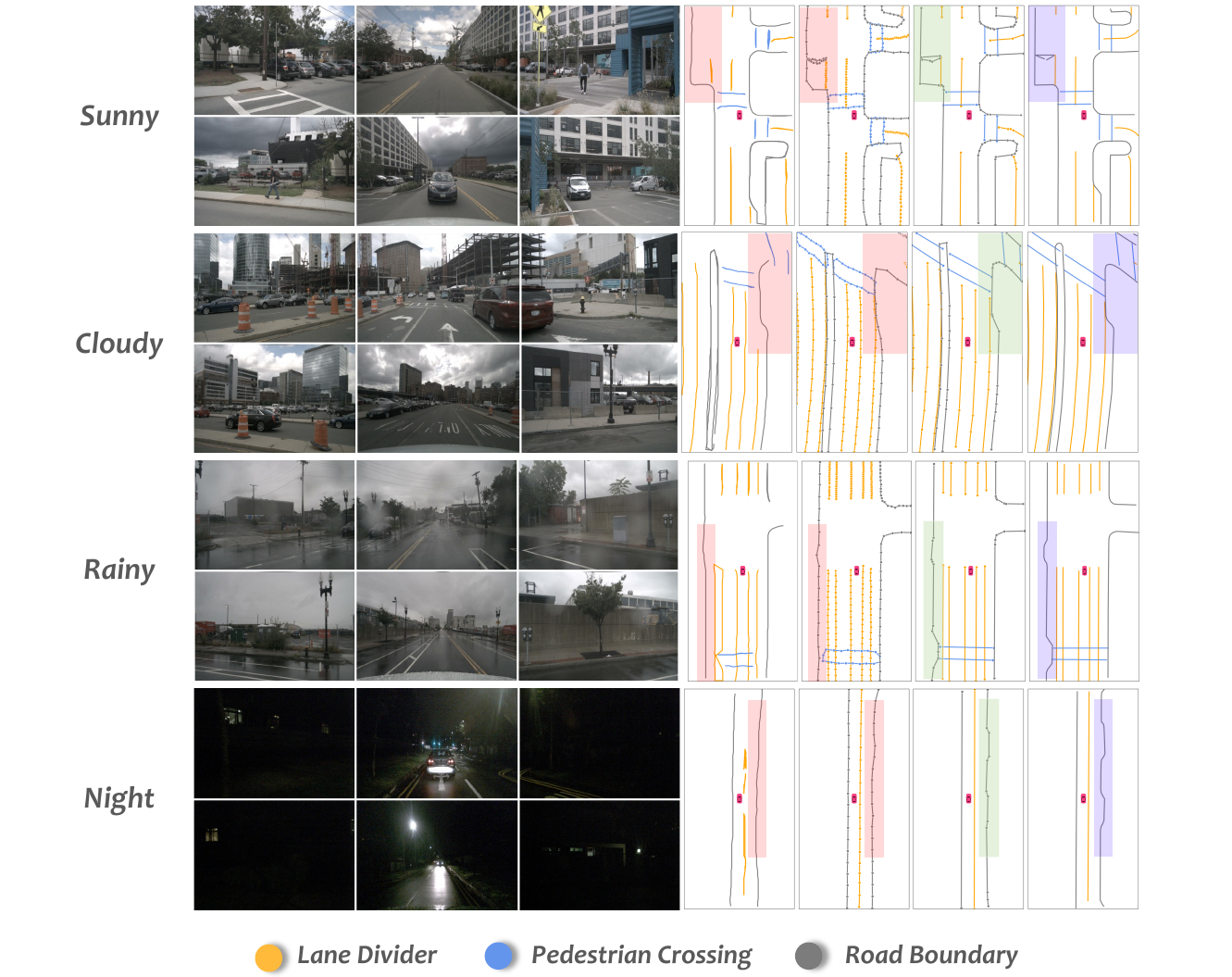} \\
	\caption{
		\textbf{Comparison with SOTAs on qualitative visualization} under different weather and lighting conditions, \ie sunny, cloudy, rainy and night. Each sub-part contains four qualitative results of HDMapNet~\cite{li2022hdmapnet}, MapTR~\cite{maptr}, \modelbf and GroundTruth respectively. The \colorbox{lightred}{red} and \colorbox{lightgreen}{green} shades show the differences and alignments with the ground truth in \colorbox{lightblue}{blue} area. Please zoom in to see more vectorized point details. 
		Note that we further provide more detailed visualization results in the supplementary material.
	}
	\label{fig:vis-main}
\end{figure*}

\minisection{Qualitative Analysis.} 
We show the qualitative comparisons with \textit{SOTAs} in Fig.~\ref{fig:vis-main} under various environment conditions.

\noindent 
\textbf{\textit{1)}} \modelit vs. \textit{HDMapNet}.
Besides avoiding complex vectorized post-processing, our pivot-based method expresses endpoints more accurately than segmentation-based ones.

\noindent 
\textbf{\textit{2)}}  \modelit vs. \textit{MapTR}.
Our approach express map shapes, \eg straight lines, rounded-corners, and right-angles, more smoothly than the method based on uniform-split polylines. Moreover, the \modelit requires fewer points for modeling.

\noindent 
\textbf{\textit{3)}}   \modelit vs. \textit{GroundTruth}.
Compared to other methods, our model is robust to various driving scenes and maintains good performance in different environments. Even at night, the map near the vehicle closely matches the ground truth.

\vspace{-0.15cm}
\minisection{Additional ablation Discussions.} 
Due to space limitation, we further provide more extensive ablation studies on encoder/decoder layer number, and predefined instance/point number in \textit{Supplementary Materials}. 

\section{Conclusion}
This paper focuses on the map construction and view the task as a direct point set prediction problem. We present an end-to-end framework for precise and compact pivot-based \hdmap construction, which embeds both geometrical and subordinate relations into map elements modeling in a unified manner.
By introducing three well-designed modules, \ie \textit{Point-to-Line Mask Module (PLM)}, \textit{Pivot Dynamic Matching (PDM)} module, and \textit{Dynamic Vectorized Sequence (DVS)} loss, the proposed \modelbf reaches \textit{SOTA} results and provides a new perspective for future research.

\clearpage

\appendix

\section*{\centering Supplementary Material}
In this supplementary material, we provide additional
details which we could not include in the main paper due to
space limitation. Specifically, we provide:
\begin{itemize}
	\setlength{\itemsep}{0pt}
	\setlength{\parsep}{0pt}
	\setlength{\parskip}{0pt}
	\item Detailed illustrations on \model design.
	\item Additional ablation studies.
	\item Extensive qualitative visualization results.
\end{itemize} 

\section{Detailed illustrations on model design}

\subsection{Code of the custom matching algorithm}

We provide the python code of the custom matching algorithm as follows. The input $cost$ refers to the distance array, where ${cost[i][j]}$ denotes the $L_1$ distance between the $i$-th point in the ground truth and the $j$-th point in the line prediction. $dp[i][j]$ denotes the lowest matching cost between the front-$i$ points in the ground truth and the front-$j$ points in the line prediction. $mem\_sort$ stores the minimum cost during
traversal to avoid unnecessary sorting. $match\_res1$ and $match\_res2$ store the temporary matching results. 

\begin{python}
def pivot_dynamic_matching(cost: np.array):
	A, B = cost.shape
	assert A >= 2 and B >= 2, \
	"A line should contain two points at least"
	if A > B: # special case
		seq_dist = 0
		for j in range(B):
			seq_dist += cost[j][j]
		combination = list(range(B))
		return seq_dist, combination      
	dp = np.ones((A, B)) * np.inf  
	mem_sort = np.ones((A, B)) * np.inf 
	match_res1 = [[] for _ in range(B)]   
	match_res2 = [[] for _ in range(B)]   
	# initialize
	for j in range(0, B-A+1):
		match_res1[j] = [0]
		mem_sort[0][j] = cost[0][0]
		if j == 0:
			dp[0][j] = cost[0][0]
	# update
	for i in range(1, A):
		for j in range(i, B-A + i+1):
			dp[i][j] = mem_sort[i-1][j-1] \
								+ cost[i][j]
			if dp[i][j] < mem_sort[i][j-1]: 
				mem_sort[i][j] = dp[i][j]
				if i < A-1: 
					match_res2[j] = match_res1[j-1] + [j]  
			else:
				mem_sort[i][j] = mem_sort[i][j-1]
				if i < A -1:
					match_res2[j] = match_res2[j-1]
		if i < A-1:
			match_res1 = match_res2.copy()
			match_res2 = [[] for _ in range(B)]  
	seq_dist =  dp[-1][-1]  
	combination = match_res1[-2] + [B-1]
	return seq_dist, combination
\end{python}

\subsection{Detailed illustrations of notations}
\label{sec:notations}
\begin{figure}[th]
	\centering
	\includegraphics[width=8cm]{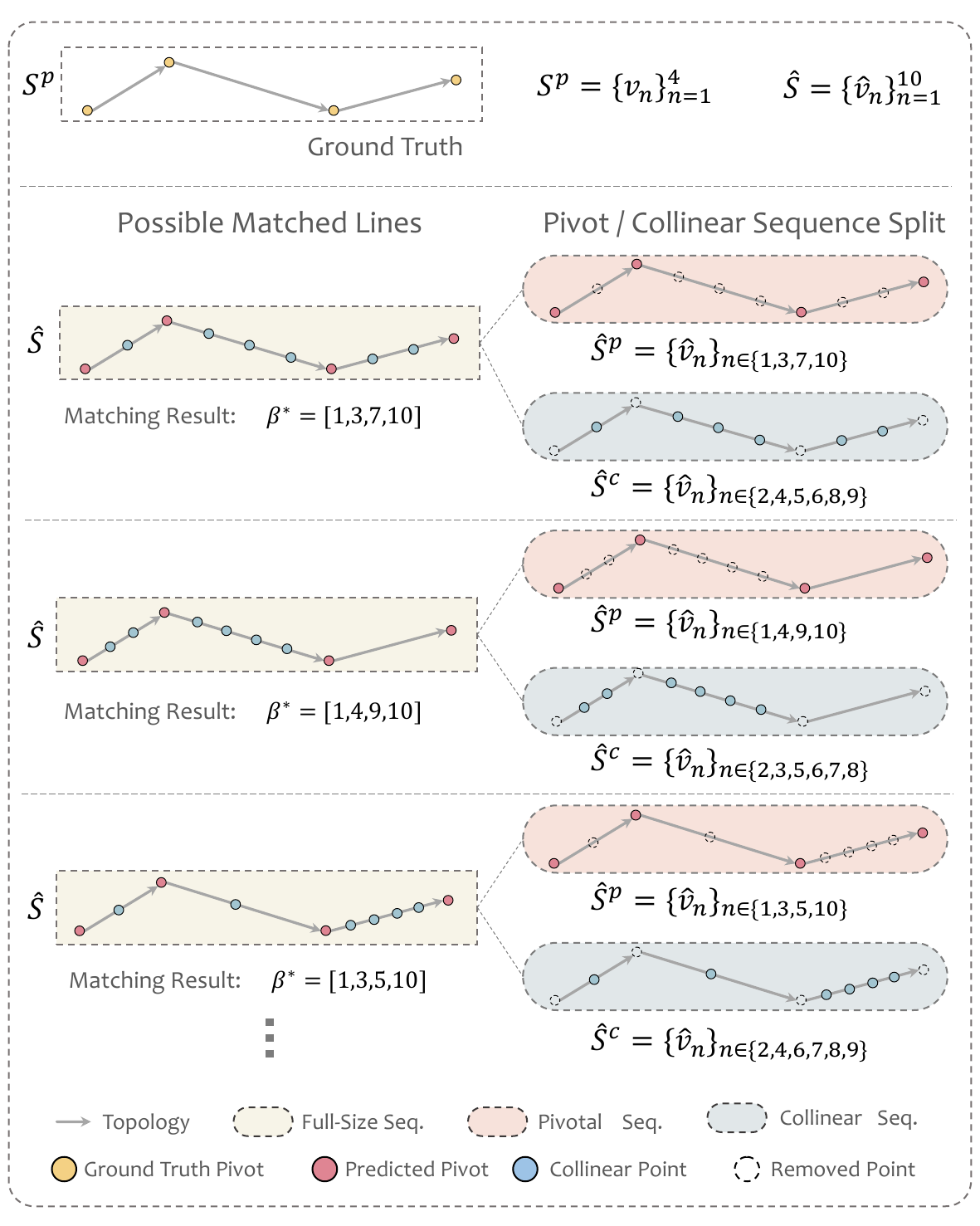} \\
	\caption{
		More illustrations of notations in pivot dynamic matching module. $\beta^*$ is the optimal $T$-combination for the predicted lines with respect to the ground truth. The figure shows the case where $T=4, N=10$. The ground truth sequence $S^p$ contains only pivot points while the predicted sequence $\hat{S}$ contains both pivot points and collinear points. $\hat{S}$ is split into a pivot sequence $\hat{S}^{p}$ and a collinear sequence $\hat{S}^{c}$ based on $\beta^*$, which is different for distinct point distribution. Predicted pivot points in $\hat{S}^{p}$ is in one-to-one correspondence to the ground truth sequence $S^p$, and $|S^p|=|\hat{S}^{p}|=T$, while the number of collinear points $|\hat{S}^{c}|$ is $N-T$. 
	}
\label{fig:illu_match}
\end{figure}

Fig.\ref{fig:illu_match} illustrates the notations of \textit{pivot dynamic matching}~(PDM) in detail. As stated in Sec.\red{3.2.3}, given a ground truth sequence $S^p=\{v_n\}_{n=1}^T$, and a predicted line $\hat{S}=\{\hat{v}_n\}_{n=1}^N$, PDM searches the optimal \textit{T}-combination $\beta^*$ with the lowest sequence matching cost among all the \textit{T}-combinations. $T$ is the length of the ground truth sequence, and $N$ is the predefined max number of points in a line prediction. For predicted lines with distinct point distribution, the optimal $\beta^*$ is different, resulting in distinct splits of pivot sequence $\hat{S}^p$ and collinear sequence $\hat{S}^c$. Fig.\ref{fig:illu_match} shows examples of $\beta^*, S^p, \hat{S}, \hat{S}^p$ and $\hat{S}^c$.

\subsection{Ground-truth pivot point generation}
\label{sec:pivotpoints}
Pivot points in the paper are defined as the points in a map element that contribute to the overall shape and typically indicate a change in direction.  
Ground-truth pivot point generation can be considered as a \textit{polyline simplification} problem, which has been studied for years~\cite{rdp,li1988algorithm,ramer1972iterative,ratschek2001robustness, vw}. Given a polyline connected by vertices, \textit{polyline simplification} aims to find a similar polyline with fewer vertices, which we call pivot points. In this paper, we choose Visvalingam-Whyatt~(VW) algorithm~\cite{vw} to generate ground-truth pivot points. Given an ordered set of points, the importance of each interior point is determined by the area of triangle formed by it and its immediate neighbors. Then the point with the smallest triangle area is obtained. If the area is below the predefined threshold, the point is removed from the set. After that, the triangle area is calculated again to find the most unimportant point. This process is repeated until no triangle area is below the threshold. 
It is worth noting that the VW algorithm~\cite{vw} is used for ground truth generation only and can be simply replaced by other pivot point generation methods like \cite{rdp,li1988algorithm,ramer1972iterative,ratschek2001robustness}.

\section{Additional ablation studies}
\label{sec:ablation}

\subsection{Impact of \bev encoder layer number}
The impact of \bev encoder layer number is evaluated in Table~\ref{tab:layer_num}. The performance of \model improves with more encoder and decoder layers. Even with single encoder and decoder layer, \model achieves satisfying performance,~\textit{i.e.,~}$36.0\%$ in mAP.

\begin{table}[ht]\scriptsize
	\resizebox{0.48\textwidth}{!}{
		\begin{tabular}{c|cccc}
			\hline
			\rowcolor{Gray}
			Layer Num. & AP$_{\textit{divider}}$ & AP$_{\textit{ped}}$ & AP$_{\textit{boundary}}$ &  mAP    \\
			\toprule
			(1, 1) &   38.2 & 33.7 & 36.0 & 36.0 \\
			(2, 4) &   45.5 & 36.9 & 42.4 & 41.6 \\
			(4, 2) &   46.3 & 38.2 & 41.8 & 42.1 \\
			\rowcolor{Gray}
			(4, 4) &  47.6 & 38.3 & 43.8 & 43.3 \\
			(6, 6) & \textbf{48.0} & \textbf{38.9} & \textbf{45.9} & \textbf{44.3} \\
			\bottomrule
		\end{tabular}
	}
\vspace{-0.15cm}
\caption{Impact of \bev encoder layer number. The gray row represents the setting we use in default.}
\label{tab:layer_num}
\end{table}

\subsection{Impact of the maximum instance number} We evaluate the effect of maximum instance number in Table~\ref{tab:Ablation-ins-num}. We define the maximum instance number based on the rule that it should be larger than the instance number in typical ground truth. In default, we choose maximum instance number of $(20, 25, 15)$ for \LD, \PC, and \RB respectively. We provide performances with other number settings in Table~\ref{tab:Ablation-ins-num}. The performance of \model get saturated when the maximum instance number is set to $(20, 25, 15)$.

\begin{table}[ht]
		\resizebox{0.48\textwidth}{!}{
			\begin{tabular}{c|cccc}
				\hline
				\rowcolor{Gray}
				Instance Num. & AP$_{\textit{divider}}$ & AP$_{\textit{ped}}$ & AP$_{\textit{boundary}}$ &  mAP    \\
				\toprule
				(15, 20, 10) &  46.5 & 37.1 & 42.8 & 42.2  \\
				\rowcolor{Gray}
				(20, 25, 15) &  \textbf{47.6} & 38.3 & \textbf{43.8} & \textbf{43.3} \\
				(30, 30, 30) &  47.6 & \textbf{38.9} & 43.2 & 43.2 \\
				\bottomrule
			\end{tabular}
		}
\vspace{-0.15cm}
\caption{Impact of the maximun instance number. The gray row represents the setting we use in default.}
\label{tab:Ablation-ins-num}
\end{table}

\subsection{Impact of the maximum pivot point number}
The impact of the maximum pivot point number is evaluated in Table~\ref{tab:sequence_len}. The maximum number of pivot points roughly represents the complexity of map elements that \model is able to model. For a certain type of map element, too small maximum number will lead to insufficient modeling capability, yet too large maximum number will increase the learning burden of the model. Therefore, an appropriate value is required for trade-off.

\begin{table}[ht]\small
	\centering
	{
		\begin{tabular}{c|ccc}
			\hline
			\rowcolor{Gray}
			Pivot Point Num. & 10 & 20 &  30 \\
			AP$_{\textit{divider}}$ & \textbf{47.6} & 47.5 & 42.6  \\
			\midrule
			\midrule
			\rowcolor{Gray}
			Pivot Point Num. & 30 & 50 & 60 \\
			AP$_{\textit{boundary}}$ & 43.8 & \textbf{45.4} & 43.2  \\
			
			\bottomrule
		\end{tabular}
	}
\vspace{-0.15cm}
\caption{Impact of the maximum pivot point number.}
\label{tab:sequence_len}
\end{table}

\vspace{-0.2cm}
 
\section{More qualitative visualization results}
\label{sec:visualization}
\begin{figure*}[h]
	\centering
	\includegraphics[width=17cm]{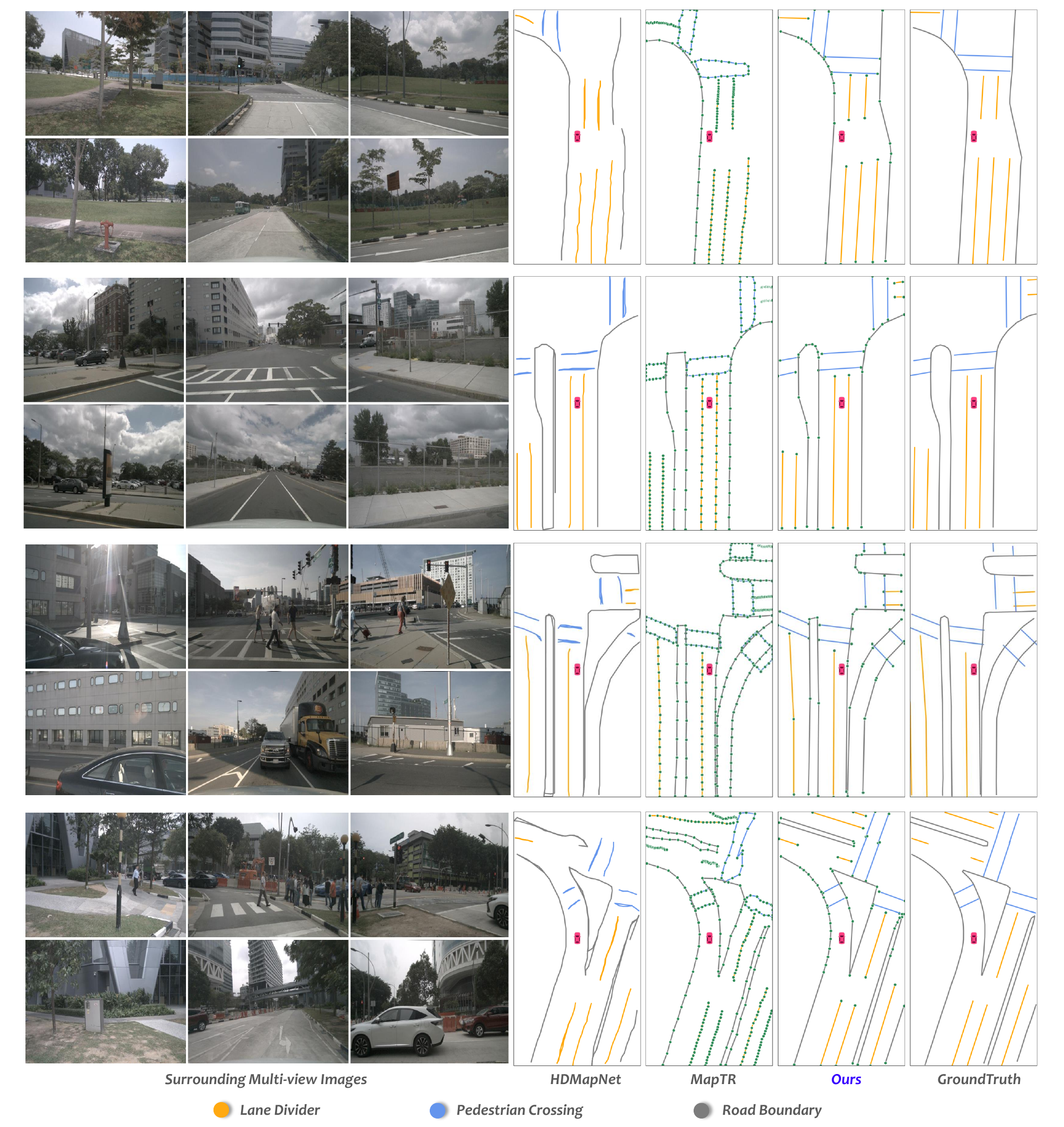} \\
	\caption{{Visualization results under the weather condition of \textit{sunny}.}}
\label{fig:sunny}
\end{figure*}

\begin{figure*}[th]
	\centering
	\includegraphics[width=17cm]{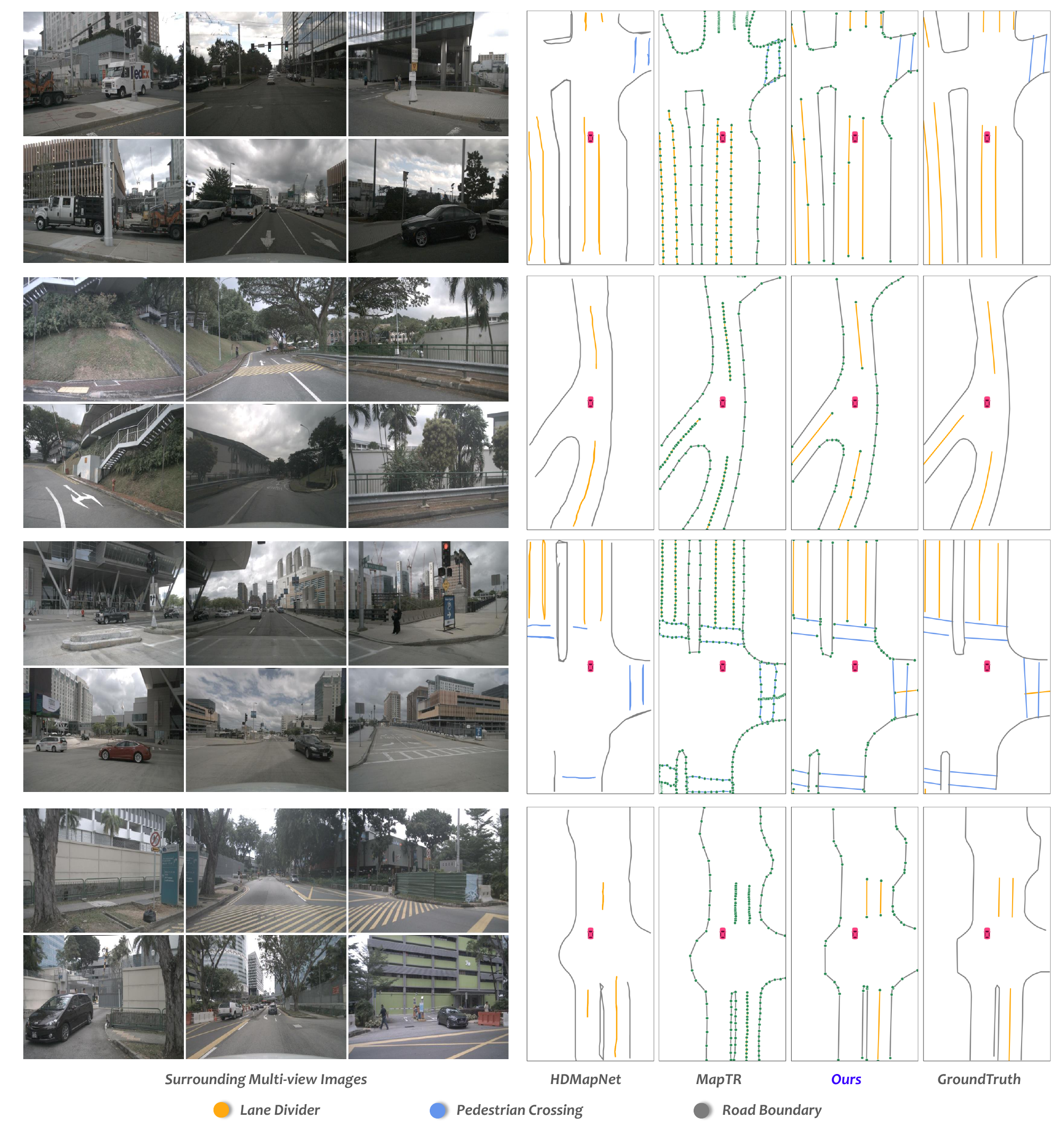} \\
	\caption{Visualization results under the weather condition of \textit{cloudy}.}
\label{fig:cloudy}
\end{figure*}

\begin{figure*}[th]
	\centering
	\includegraphics[width=17cm]{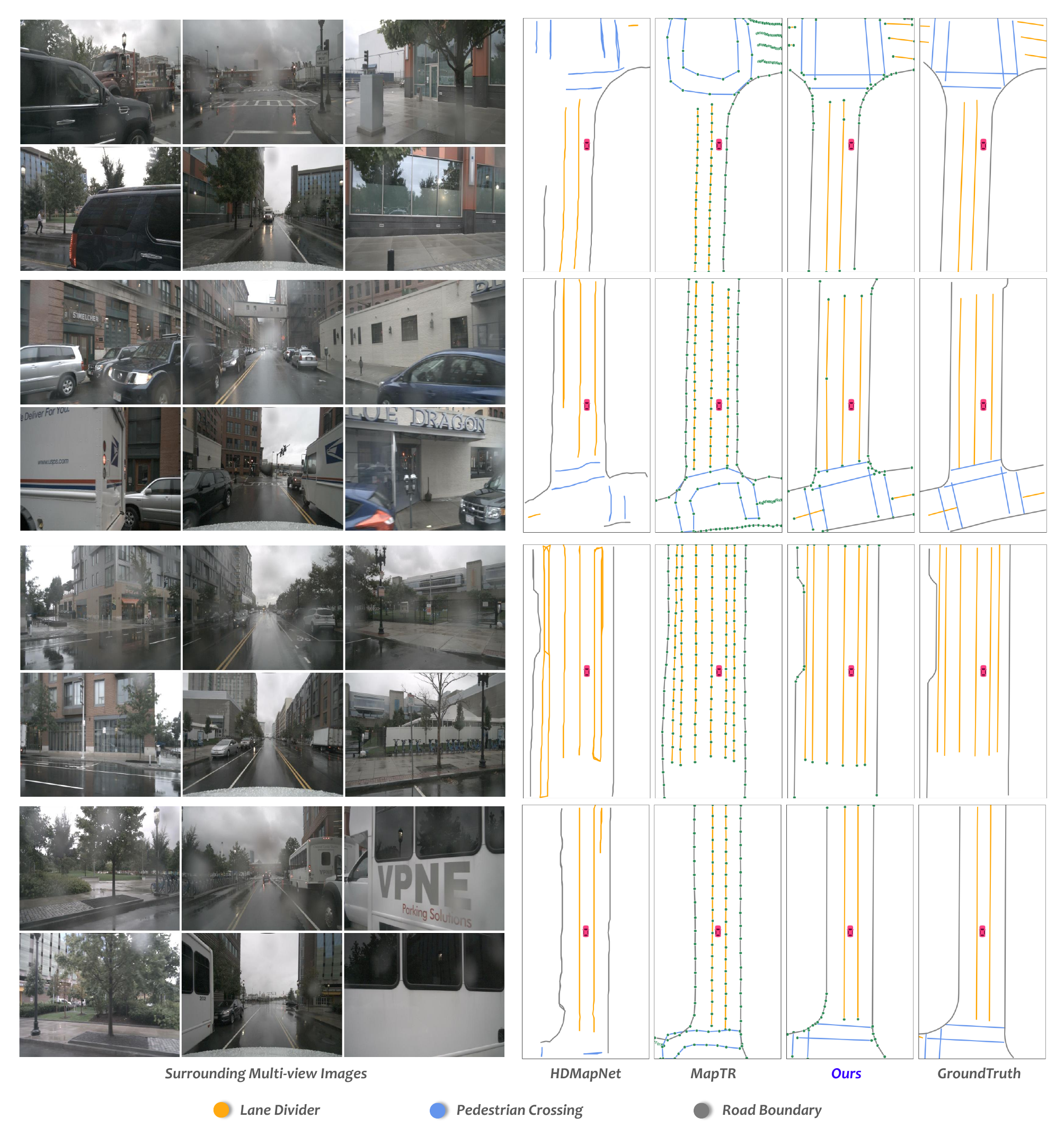} \\
	\caption{Visualization results under the weather condition of \textit{rainy}.}
\label{fig:rainy}
\end{figure*}

\begin{figure*}[th]
	\centering
	\includegraphics[width=17cm]{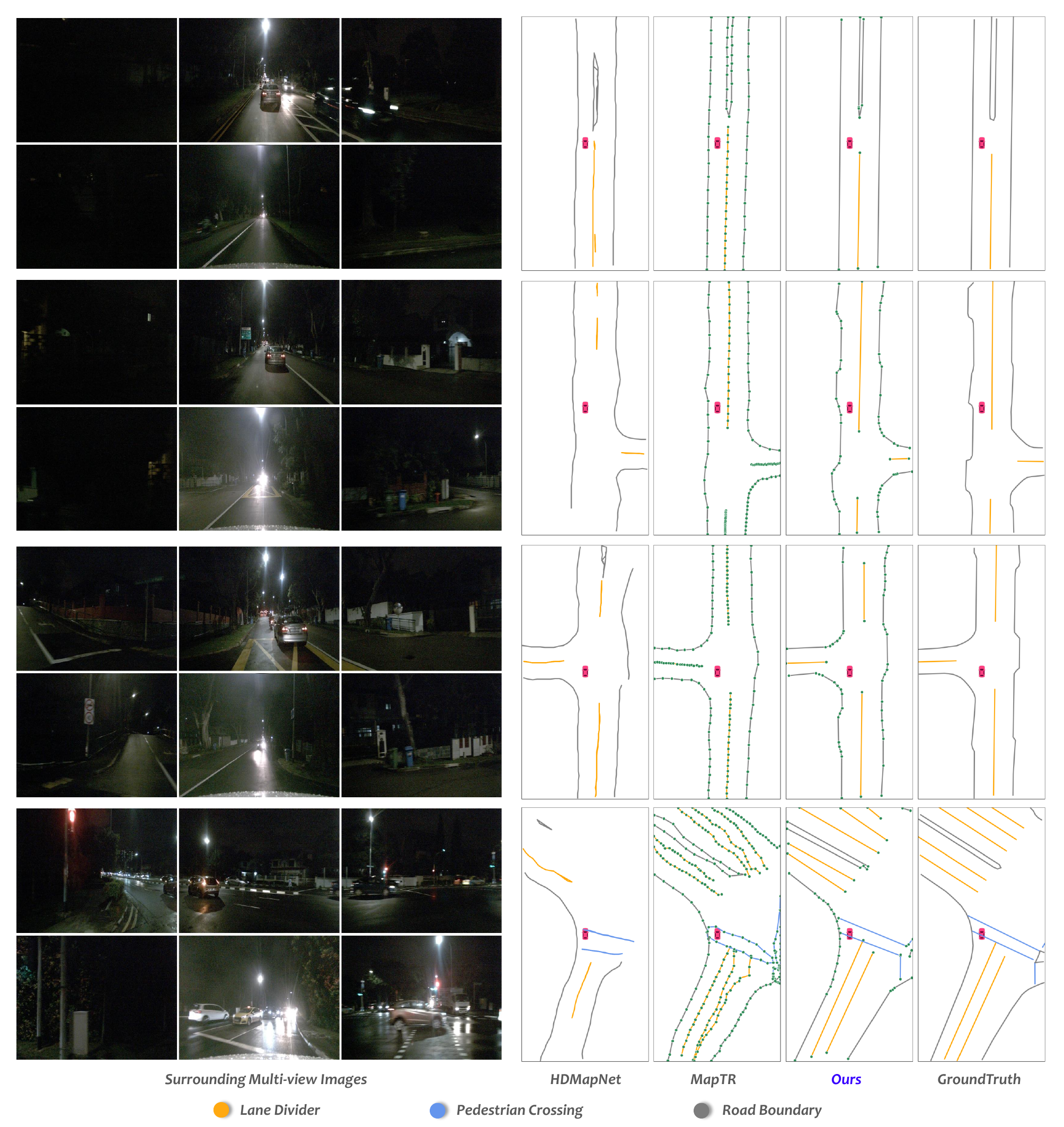} \\
	\caption{Visualization results under the lighting condition of \textit{nighttime}.}
\label{fig:night}
\end{figure*}

Fig.\ref{fig:sunny}-\ref{fig:night} provide extensive visualization results for comparison with SOTA approaches~\cite{li2022hdmapnet,maptr} on various environmental conditions.
The visualization of HDMapNet~\cite{li2022hdmapnet} is reproduced with its public code, and that of MapTR~\cite{maptr} is generated by its released model checkpoint. 
Even in challenging road scenarios like intersections and dense roads, \model produces accurate yet compact representations.
 
We visualize the predictions of the \model with Swin-Tiny~\cite{liu2021swin} backbone on nuScenes.
The predictions on different environmental conditions are presented. Even at the most challenging night scenario, the predictions near the vehicle closely match the ground truth~(see Fig.~\ref{fig:night}). Seen from visualization, the modeling of \model is flexible and can describe map elements of arbitrary shape, including line segments, curves, and combinations of them.

\clearpage

\clearpage

{\small
\bibliographystyle{ieee_fullname}
\bibliography{egbib}
}

\end{document}